\documentclass[conference,compsoc]{IEEEtran}
\ifCLASSOPTIONcompsoc
  \usepackage[nocompress]{cite}
\else
  \usepackage{cite}
\fi
\ifCLASSINFOpdf
\else
\fi
\usepackage[ruled,vlined,linesnumbered]{algorithm2e}

\usepackage{graphicx}

\begin{document}
%
\title{Credit Card Fraud Detection: A Deep Learning Approach}

\author{\IEEEauthorblockN{Sourav Verma}
\IEEEauthorblockA{ABV-Indian Institute of \\Information Technology and Management\\Gwalior-474 010\\
Email: ipg\_2013108@iiitm.ac.in}
\and
\IEEEauthorblockN{Joydip Dhar}
\IEEEauthorblockA{ABV-Indian Institute of \\Information Technology and Management\\Gwalior-474 010\\
Email: jdhar@iiitm.ac.in}}


%


\maketitle

\begin{abstract}
Credit card is one of the most extensive method of installment for both online and offline mode of payment for electronic transactions in recent times. credit cards invention has provided significant ease in electronic transactions. However, it has also provided new fraud opportunities for criminals, which results in increased fraud rates. Substantial amount of money have been lost by many institutions and individuals due to fraudulent credit card transactions. Adapting improved and dynamic fraud recognition frameworks thus became essential for all credit card distributing banks to mitigate their losses.
\\  \par In fact, the problem of fraudulent credit card transactions implicates a number of relevant real-time challenges, namely: Concept drift, Class imbalance, and Verification latency. However, the vast majority of current systems are based on artificial intelligence (AI), Fuzzy logic, Machine Learning, Data mining, Genetic Algorithms, and so on,  rely on assumptions that hardly address all the relevant challenges of fraud-detection system (FDS). 
\\ \par This paper aims to understand \& implement Deep Learning algorithms in order to obtain a high fraud coverage with very low false positive rate. Also it aims to implement an auto-encoder as an unsupervised (semi-supervised) method of learning common patterns.
\end{abstract}


%
\IEEEpeerreviewmaketitle

\section{Introduction}
Fraud refers to the intentional illegal exploitation of a system which results in loss of an oblivious entity. Credit card fraud can be defined as illegal use of both online and offline mode of credit card information for electronic transactions. Credit card fraud involves the exploitation of credit card systems which results in the loss of financial resources, the most prominent being monetary although other damages such as loss of integrity and authenticity are possible. Fraud, waste, and abuse in many financial systems are estimated to result in billions of US dollars annually, thus has became a primary concern for financial institutions around the globe. \\ \par Furthermore, the rapid growth of the internet has exposed credit card systems to diverse fraudsters using different mechanisms to exploit financial systems. This provided an explode in attack patterns which rendered the once effective Artificial intelligent (AI), Machine Learning and Case-based fraud detection solutions no more effective as the computational complexity increases with each new detected fraud. More seriously, there is a higher tendency for first time frauds going undetected. The Case-based detection methods are also slow as a successful exploit could multiply if the solution took time to be integrated into the system. This problem can only be addressed with a refined and dynamic techniques capable of adapting to rapidly evolving fraudulent patterns. \\ \par  Also of concern to credit card fraud detection solutions is, the \textit{recognition strength} that indicates a Fraud detector's ability to correctly identify both known and novel frauds. This is usually a direct function of how much fraud samples there are to model a solution. The emergence of Deep Learning algorithms has provided credit card fraud detection experts with verse amount of features and high dimentionality of data that will enhance the detection models. Such solutions that use Deep Learning algorithms to model, offer more efficient and adaptive solutions. \\ \par A complete credit card fraud detection model thus,\\ \\ Must have the following properties: 
\begin{enumerate}
	
	\item \textit{It must be Adaptive:} This refers to following abilities:
	\begin{itemize}	
		\item  Ability to detect fraudulent patterns in a matter of moment (i.e quickly). This is also referred to as its \textit{alertness}.
		\item Ability to detect \textit{first time} fraudulent patterns with high accuracy and low false positive rate.	
	\end{itemize}
	\item \textit{It must be Dynamic:} This refers to following abilities:
	\begin{itemize}
		\item Ability to detect all \textit{new instances} of fraudulent activities with rapid alteration of false patterns.
	\end{itemize}
	\item It should be able to identify the fraudulent patterns accurately i.e. the true positive rate should be high.
	\item It should be able to detect the frauds quickly i.e time complexity of the system should be low.
	\item It should not predict a legit transaction as fraud i.e. the false positive rate should be low.
	
\end{enumerate}
Also must address following relevant challenges, namely:
\begin{enumerate}
	\item \textbf{Concept drift:} As customers' habit changes, fraudsters change their approaches over time.
	\item \textbf{Class imbalance:} More legit transactions as compare to frauds (less than 0.5\%).
	\item \textbf{Verification latency:} Only a small set of transactions are timely checked by the authorities.
\end{enumerate}

\section{Related Works}
\subsection{Frauds and Fraudulent pattern detection}
Fraud is a synonym for illicit use of a system to get some benefits, usually resulting in loss to another person. Frauds are as diverse as fraudulent patterns. Credit card fraud is fraud within the financial industry that usually cause monetary losses. The financial industries have been the principal victims of fraudulent activities in recent times. According to \cite{dal2017credit}, substantial amount of money have been lost to insurance fraud. The growth of internet uses have made it easier for fraudsters to divulge and connect in order to target financial institutions from a distance, making it more diverse in nature. This further entangles the threats to  credit card security systems, thus fraud detection and prevention are important concern to all financial institutions.\\ \par By many estimates, approximately 10\% of insurance company payments are for fraudulent claims, and the global sum of these fraudulent payments, amounts to substantial amount of money. \\ \par Fraud detection refers to mechanisms to \textit{detect} frauds when fraudulent activities occur, while Fraud prevention refers to all measures put in place to \textit{prevent} frauds from happening,  \cite{bolton2002statistical}. A necessary requirement for fraud preventive systems is their precision (i.e. \textit{predictive measure}). Much concern is given to improving the precision of such systems.  Fraud detection systems, conversely, need to adapt to the dynamism of threats. Hence in addition to possible predictiveness, Fraud detection systems need to be \textit{adaptive} to the fraudulent patterns. A related concern usually classified under possible predictiveness is the time complexity of the system to detect fraudulent transactions. Certain systems require \textit{near real-time} alertness on dubious transactions.

\subsection{Outlier detection}
An outlier is an observation that deviates so much from other observations as to evoke suspicion that it was generated by a different mechanism \cite{hung1999parallel}. \\ \par Unsupervised learning approaches are used to this model. Usually, the result of unsupervised learning is a new explanation or representation of the original observation data, which then lead to improved future responses or decisions. Unsupervised learning methods do not need prior knowledge of fraudulent and genuine transactions (i.e. labels) in historical databases, instead detect alteration in behavior or unusual transactions. These methods model a baseline distribution that represents normal behavior and then detect observations that show deviation from this norm. Outliers are a basic form of non-standard observation that can be used for fraud detection. In supervised methods, models are trained to discriminate between fraudulent and non-fraudulent behavior so that new observations can be assigned to different classes. Supervised methods require accurate identification of fraudulent transactions in historical databases and can only be used to detect frauds of a type that have previously occurred. An advantage of using unsupervised methods over supervised methods is that previously undiscovered types of fraud may also be detected. Supervised methods are merely trained to discriminate between legit transactions and previously known fraud. \par \cite{bolton2002statistical} proposed unsupervised credit card fraud detection techniques, using behavioral outlier detection techniques. Anomalous spending behavior and frequency of transactions can be identified as outliers, which could be possible fraud cases.

\subsection{Rule-based fraud detection}
Rule-based methods consist all known fraudulent characteristics and use them to model the fraud detection system (FDS). They are classified as Supervised learning methods as they use previously known fraud to detect similar patterns. Such methods classify transactions using rules made out based on previously detected fraudulent transactions. The process used to adopt such models to evolving threats is manual, and thus such methods are not recommended for persistent threats of these days. An example of such methods include BAYES, RIPPER etc.\\ \par  Although according to \cite{kou2004survey}, Rule-based fraud analysis can be very tedious to administer because the proper layout of such rules require detailed, onerous, and prolonged programming for each credible fraud instance. The dynamic emergence of multiple new fraud types, demands that these rules be constantly adapted to include existing, emerging, and future fraud options. Moreover, it also presents a major hurdle to \textit{scalability}. The more data the system must process, the more severe is the performance descents.

\subsection{Statistical fraud detection} 
Statistical methods have been used to classify and detect frauds. The transactional data is known to follow a statistical distribution, and thus, transactional data points, that fall out of the normal distribution are considered dubious. Such methods include Linear Discriminant Analysis (LDA) and Logistic Regression \cite{bolton2002statistical}. Statistical methods can either be supervised or unsupervised. Supervised methods use known fraudulent cases to model the detector system. \par A natural problem with the Statistical method is determining the most relevant distribution to fit a data set (best fit), and with increased dimension of the data, it becomes more difficult to approximate the distribution, \cite{tan2005introduction}.

\subsection{Machine Learning based detection}
Machine learning (ML) is the science of getting computers to act without being explicitly programmed. In the past decade, machine learning has given us self-driving cars, practical speech recognition, effective web search, and a vastly improved understanding of the human genome.

ML evolved primarily from Artificial Intelligence (AI) and Soft computing, and also from other fields including applied mathematics, pattern recognition and computational learning theory \cite{adewumi2016survey}. ML algorithms are mostly used to handle problems involving automatic data classification \cite{bergholz2008improved}. ML algorithms are capable of analyzing data and searching for hidden patterns in data. According to \cite{ayodele2010types} ML algorithms aims to predict patterns from data based on learned experiences. ML algorithms are divided into different classes, namely: supervised learning, unsupervised learning, semi-supervised learning, reinforcement learning, transduction and learning to learn \cite{ayodele2010types}. Most of the proposed credit card fraud detection techniques are based on supervised learning and few are based on semi-supervised learning. Some of these techniques are discussed next:

\subsubsection{Hidden Markov Model}
\cite{srivastava2008credit} proposed a technique based on Hidden Markov Model (HMM). In the study, authors used HMM to model a sequence of credit card transactions and divide the transactions into three price ranges (clusters): low (l), medium (m), and high (h). The type of each transaction is linked to the line of business of the corresponding merchant. Afterwards, determine the three probability matrices so that representation of the HMM is complete. These three model parameters are determined in a training phase using the Baum-Welch algorithm \cite{khan2014credit}. Authors considered the special case of fully connected HMM in which every state of the model can be reached in a single step from every other state. \\
An HMM is initially trained with the normal behaviour of the cardholder. Thereafter, authors constructed sequences from training data-set and trained the model. In the testing and validation phase, if an incoming credit card transaction is not accepted by the trained HMM with sufficiently high probability, it is considered to be fraudulent. authors also claim to minimize the true positive rates.  \\ \par Another proposed technique based on HMM is proposed by \cite{khan2014credit}. In the study also, authors used HMM to model a sequence of credit card transactions and used K-mean clustering algorithm to cluster the transactions into three price ranges (clusters): low (l), medium (m), and high (h), as proposed by \cite{srivastava2008credit} Afterwards, incoming transactions were tested into the trained model and authorized if accepted with sufficiently high probability. Otherwise, transaction will be terminated and IP address of the merchant to be defrauded will be traced using HMM. A notification will be sent to both the merchant system’s administrator and cardholder via mobile communications. Also as per the authors, the HMM was trained with Baum-Welch algorithm.

\subsubsection{Support vector machines (SVM) based techniques}
\cite{csahin2011detecting} performed a comparative study between SVM and decision tree based credit card fraud detection system (FDS). Firstly, authors divided the data-set used into three groups with the ratio of fraudulent transactions to the legit ones in 1:1, 1:4, 1:9 respectively, during the implementation. As usual training to testing data-set was divided by 70\% to 30\%. Authors used four kernels for SVM in the setup. Also they developed seven SVM-based and decision tree based models and tested each of them. Results from experiments revealed that the Decision tree based model outperformed SVM model. The models achieved classification accuracy between the range of 83.02 to 94.76\%.

\subsubsection{Frequent item-set mining}
\cite{seeja2014fraudminer} proposed a technique based on frequent item-set mining, called FraudMiner. Authors separated each customer's transaction from the whole transactional database and from each customer's transactions again separated their legit and fraud transactions. Afterwards, applied Apriori algorithm to both the sets of legit and fraud transactions on each customer's transactions which returns a set of frequent item-sets of both legit and fraud transactions. for testing, authors propose a matching algorithm which spans the legit and fraudulent pattern databases of each customers for a match with the incoming transaction to detect fraud. If a convenient match is found with legit pattern of the corresponding customer, it matches, otherwise not. The experimental result shows that the FraudMiner outperformed the existing solutions eg. SVM, Naive Bias (NB), KNN etc.

\subsubsection{Ensemble based technique}
\cite{zareapoor2015application} proposed a credit card
fraud detection model based on bagging ensemble classifier. The primary objective of study was to compare the performance of three different advanced data mining techniques namely: SVM, NB and KNN to bagging ensemble classifier based on decision tree. To counter class imbalance problem authors divided the data-set used into four groups with fraud rates approximately 20\%, 15\%, 10\%, 3\% respectively also authors used 10 fold cross validation technique. To compare the results authors weigh the performance of SVM, NB and KNN and compared with the result obtained by bagging ensemble classifier. The experimental result revealed that bagging ensemble classifier achieved better fraud detection rate and an improved false positive rate. \\ \par
\cite{fadaei2017ensemble} proposed a credit card fraud detection techniques based on Ensemble classification and extended feature selection. Authors considered both the feature selection and the prediction (decision) cost for accuracy enhancement of the FDS. After selecting best features using an extended wrapper method, an ensemble classification is performed. Authors performed the ensemble classification using cost sensitive decision tree in a decision forest framework. The experimental result revealed that considering the F-measure as the evaluation metric, the proposed approach achieves 1.8 to 2.4\% performance improvement compared to the other classifiers.

\subsection{Nature inspired (NI) based techniques}
Nature inspired based methods refers to algorithms inspired by nature's problem solving ability \cite{rozenberg2011handbook}. In other words, Nature is
the source of inspiration to Nature Inspired algorithms. For example, Ant Colony Optimization is inspired by the methods used by ants to seek for pathways between their colony and a food source, Bat algorithm was inspired by the echolocation behaviour of micro-bats, with varying pulse rates of emission and loudness and Genetic Algorithm is inspired by the process of natural selection, that belongs to the larger class of evolutionary algorithms (EA). \cite{mitchell1998introduction}. NI algorithms are designed to handle complex real world classification and optimization related problems, such as timetabling problem, travelling salesman problems (TSP) and hostel allocation problems \cite{rozenberg2011handbook}. Generally, NI algorithms are used for global optimization. Some NI-based techniques used to provide solution to credit card fraud detection are discussed next.

\subsubsection{Genetic algorithm (GA) based techniques}
\cite{patel2013credit} proposed a GA-based credit card FDS with the intent of detecting the fraud with minimum false positive rate. Instead of maximizing the correctly classified transactions authors prescribed an
objective function with variable misclassification cost. The objective function intents at minimizing false positive rate. During classification, authors extracted credit card transaction from the database and standardize the data. Afterwords, calculated critical values for each transaction present in the database. Authors also extracted the frequent item-set for credit card usage, indigence, location where the credit card was used, balance on the account linked to credit card, average spending pattern of the credit cardholder from each transaction. Furthermore, authors used Genetic Algorithm (GA) to generate new critical values. Finally, the new critical values were then used for classification. 

\subsubsection{Artificial neural network (ANN) based techniques}
\cite{awlla2017hybrid} used Simulated annealing (SA) and
Back-propagation algorithm (BPA) for Feed-forward
Neural Network (FFNN) to develop a credit card FDS with the intent of hybridizing SA and BPA for FFNN, which can join the symbolic global searching capability of SA with the precise local searching element of back-propagation FFNNs to improve the initial weights of a neural network toward getting a better result for detection fraud. Authors suggest of identifying fraud and legit transactions based on following critical values, namely: credit card usage frequency, number of locations of credit card usage, average credit card indigence and credit card book balance. \par Authors randomly initialized the weights of FFNN and evaluated weights using SA, following a temperature annealing schedule with the algorithm. While first temperature value is less than or equal to the minimum error authors selected the best solution. Furthermore, for training, authors initialize the parameters of BP learning algorithm. While the threshold epoch not reached, authors update the weights of BP to minimize the error with training data. Finally, Authors assess the execution of classification with test data to validate the study. The experimental result revealed that BPFFNN with applied SA yielded better false positive rate as compare to the simple BPFFNN algorithm.

\section{Proposed Methodology}
To address the major concerns of a Fraud detection system, the following infrastructural design choices were made.

\subsection{Deep Learning Model}
A \textbf{deep learning} computation model will be used to model FDS. Deep learning models are appropriate here as they serve well for data sets having large amount of data with large no. of features. The ability of such models to learn feature hierarchy composing lower level features into higher level abstractions influenced its choice here. It has the potential to discover sophisticated patterns in large data sets through its self-adjusting back-propagation algorithm.

Furthermore, Deep Learning models address some of the big challenges posed by Big Data computation and analysis, thereby providing efficient means to the use of Big Data.

For the purpose of this task, two Deep Learning models are used to model the credit-card fraud data and to predict fraudulent transactions. The first model will be a \textit{Multi-layer feed-forward neural network system}. It is built based on the neuron units. The model works by feeding the input data into the first layer. Subsequent layers learn more concrete features from previous layers through \textit{non-linear} transformations. 
\\ 

\textbf{Experiment I: Multi-layer feed-forward supervised learning} 
\\ \\
This experiment involves tuning a feed-forward deep learning network to
model financial data. Parameters such as activation function, number of
epochs, hidden layers size will be adjusted until a suitable model is reached.
The aim of the experiment is to find recommendations for Fraud detection
system designs. \par
Choices for activation function include: The \textit{Tanh function \& Logistic sigmoid function} with \textit{min-max or z-score normalization}. After Using validation set to determine the data standardization approach and the best fit activation functions, According to experiments, It's found that Tanh performs better than Logistic sigmoid, when using it with z-score normalization. Hence, choosing tanh followed by z-score is the best option. the Epoch numbers will be increased by factors of 10 (1, 10, 100 etc.). Hidden layers will vary between 1 to 50. The \textit{mean-squared error (MSE) function or reconstruction error} is used as the \textit{loss function}.

\par The multi-layer feed-forward Deep Learning model is used to conduct supervised learning with the training data set, which is the majority of the data, splitted in the ratio of 3:1. Design parameters will be varied to determine a recommended design for the FDS on the data set.

\subsection{Anomaly Detection}\label{sec4.2.2}
The anomaly detection model used here is \textit{Deep Auto-encoders (DAE)} to detect fraudulent pattern in the data-set which is an unsupervised model. \par Unsupervised means the model will be trained for both fraud and non-fraud data without feeding the labels. Since the class imbalance is very high in the credit cards, It is expected from the model to learn and memorize the patterns of legit ones after the unsupervised training, and should be able to give a credible score for any transaction as being an outlier. And this unsupervised training would be quite handy in practice especially when we don't have enough labeled data set. Deep Auto-encoders can be used to pre-train the model before a supervised training. \\

\textbf{Experiment II: Deep Auto-encoders for detecting fraudulent patterns} 
\\ \\
This experiment involves designing a Deep Auto-encoder to detect anomaly in the credit card data-set. The Deep Auto-encoder learns the pattern in the data-set through non-linear transformations of layers. \par To test for anomaly, it reconstructs the test data, anomalous data will deviate a lot from the legit ones and thus will have high error. A deep learning auto-encoder will be trained on 75\% of the data set, the remaining 25\% will be used to test for the model’s predictions. The \textit{mean-squared error (MSE) function or reconstruction error function} is used as the \textit{loss function}.
\\ \\
\textit{Reconstruction Error Function:}
\begin{equation}
    L(x, x') = \left|| x - x' \right||^2
\end{equation}

\textbf{Experiment III: Optimization} 
\\ \\
This experiment involves optimizing the performance of the proposed deep learning fraud detection model. For optimizing the performance of the FDS, \textit{Nature Inspired Bat Algorithm} is used. The aim is to mitigate the training cost and complexity of the FDS. Also to enhance the overall performance of the model.
\\

\textbf{Experiment IV: Comparison with different Scikit learn algorithms} 
\\ \\
This experiment involves comparing different \textit{Scikit-learn} methods/algorithms to classify fraudulent patterns. These comparisons are done using different performance measures e.g.: \textit{AUC scores, confusion matrices and precision-recall curves}.

\section{Implementation \& Execution}
We trained and tested our proposed FDS using Kaggle's Credit Card Fraud Detection data-set. The summary of data-set:
\\ 
\begin{center}
    \begin{tabular}{ |p{4cm}||p{3cm}| }
     \hline
     \multicolumn{2}{|c|}{Kaggle: Credit Card Fraud Detection Data-set} \\
     \hline
     Data Set Characteristics & Multivariate\\
     \hline
     Number of Attributes & 31\\
     \hline
     Number of Instances & 284807\\
     \hline
     Attribute Characteristics & Categorical, Float64\\
     \hline
    \end{tabular}
\end{center}

\subsection{Data Standardization \& Activation Function}
Two types of data standardization function is considered here: \textit{z-score \& min-max normalization}. 1) z-score normalization will normalize every column such that the resultant columns will have mean of zero and standardization of ones. And this will be a good choice if we are using \textit{Tanh} activation function. This will output values on both sides of zero. Furthermore, Tanh activation function will leave values that are too extreme to still keep some outliers left after the normalization process. This might be useful to detect some extremeness in this case.\par 2) min-max normalization will assure all values to be in the range [0, 1]. min-max is the default scaling approach if we are using sigmoid as our output activation function. \\ 

\textit{Z-Score Standardization Function:}
\begin{equation}
    z-score = \frac{x - \mu}{\sigma}
\end{equation}

\textit{Tanh Activation Function:}
\begin{equation}
    tanh = \frac{e^{2x} - 1}{e^{2x} + 1}
\end{equation}

\subsection{Modeling Auto-encoder as unsupervised learning}

\begin{itemize}

  \item \textbf{Parameters:}
    \begin{itemize}
      \item learning rate = 0.01
      \item training epochs = 60 (Optimal)
      \item batch size = 256
      \item display step = 1
    \end{itemize}
    
 \item \textbf{Network Parameters:}
    \begin{itemize}
      \item n\_hidden\_1 = 15
      \item n\_hidden\_2 = 15
    \end{itemize}
    
 \item \textbf{For FC layers:}
    \begin{itemize}
      \item hidden\_size = 4 (Best hidden size based on validation)
      \item output\_size = 2 (classes: 1 \& 0)
    \end{itemize}
    
\end{itemize}

\subsection{Optimization}
Optimization is done using \textit{Nature Inspired Bat Algorithm}.

\subsubsection{Binary Bat Algorithm}
The binary bat algorithm has been inspired by the echolocation behaviour of bats \cite{yang2014nature}. The characteristics of bats for finding its pray are being used in this algorithm. Bats tend to decrease the loudness and increase the rate of emitting ultrasonic waves, when they chase pray.\par
In binary bat algorithm each artificial bat has a position vector, a velocity vector and a frequency vector. The position of the bats in binary bat algorithm is either 0 or 1. The velocity can be updated using the following equations:
\begin{equation}
    V_{i}(t+1) = V_{i}(t) + (X_{i}(t) - X^*)F_{i}
\end{equation}
Where Vi , Xi and Fi are the velocity, position and frequency of ith bat. X* is the current global best location. 
\par The frequency of the ith bat can be updated using the following formula:
\begin{equation}
    F_{i} = F_{min} + (F_{max} - F_{min})\beta
\end{equation}
Where $F_{min}$ is the minimum frequency and $F_{max}$ is the maximum frequency. $\beta$ represents a random number which lies between 0 and 1. \par
The position of bats can be updated based on following function:
\begin{equation}
    X_{i}(t+1) = X_{i}(t) + V_{i}(t)
\end{equation}
The loudness and pulse rate of binary bat algorithm is A and r. These two variables can be updated as follows:
\begin{equation}
    A_{i}(t+1) = \alpha A_{i}(t)
\end{equation}
\begin{equation}
    r_{i}(t+1) =  r_{i}(0)[1 - \exp(-\gamma t)]
\end{equation}
Where $\alpha$ and $\gamma$ are constants. The loudness and the pulse rate are updated when we optimize the new solutions to ensure that the bats are moving toward the best solutions.

\subsubsection{Optimization using Binary Bat Algorithm}
By using \textit{Nature Inspired Bat Algorithm} for feature selection process, It is found that some features can be dropped to mitigate the training cost and the complexity of the entire system. Also it enhances the test AUC score significantly.\\
Features dropped were: \par
'V28', 'V27', 'V26', 'V25', 'V24', 'V23', 'V22', 'V20', 'V15', 'V13', 'V8'

 
\begin{algorithm}[]
\DontPrintSemicolon
\SetAlgoLined
\BlankLine
Initialize the bat's position and velocity, $X_{i}$ and $V_{i}$ (i = 1, 2, 3, .. n)
\BlankLine
Initialize frequency ($F_{i}$), pulse rate ($r_{i}$) and loudness ($A_{i}$)
\BlankLine
\While{$iteration < max\_iteration$}{
    Generate new solutions by adjusting frequencies\;
    Update velocity and locations for each solution\;
    \If{$rand > r_{i}$}{
        Select a solution among the best solution\;
        Generate a local solution among the selected best\;
    }
    Generate a new solution by flying (around) randomly\;
    \If{$rand < A_{i} and F(X_{i}) < F(X^*)$}{
        Accept the new solution\;
        increase $r_{i}$ and reduce $A_{i}$\;
    }
    Rank the bats and find current best X*\;
}
 
\caption{Pseudo code for Binary Bat Algorithm:}
\end{algorithm}

\section{Result \& Analysis}
This section presents the outcomes of the experiments designed to provide Fraud detection systems. \\

\subsection{Performance Measures}
The following standard evaluation measures were used in describing the predictive Fraud detection models:
\begin{enumerate}
    \item \textbf{Mean Square Error (MSE):} It is used to assess the quality of a predictor
    models. It takes values in the range [0, 1]. The goal is to minimize MSE value.
    A ideal model will have MSE value = 0.
    
    \item \textbf{Accuracy:} Accuracy is the simplest performance measure that refers to the closeness of a measured value to a standard or known value. Mathematically, it is the ratio between the number of correct predictions and the total number of predictions.
    \begin{equation}
        Accuracy = \frac{\# Correct}{\# Predictions}
    \end{equation}
    
    \item \textbf{Confusion Matrix:} A confusion matrix (or confusion table) is a table that is often used to describe the performance of a prediction model on a set of test data for which the true values are known. In other words, it is a matrix between \textit{Real Classes} and \textit{Predicted Classes}.\par
    Confusion Matrix has following terms:
    \begin{itemize}
        \item \textbf{True Positive (TP):} Means no of positive cases which are predicted positive.
        \item \textbf{False Positive (FP):} Means no of negative cases which are predicted positive.
        \item \textbf{True Negative (TN):} Means no of negative cases which are predicted negative.
        \item \textbf{False Negative (FN):} Means no of positive cases which are predicted negative.
    \end{itemize}
    
    \item \textbf{Precision \& Recall (PR):} Precision is a measure of classifier's exactness. Whereas, Recall is a measure of classifier's exactness. \par
    In other words, Precision refers to how many selected items are relevant? Whereas, Recall refers to how many relevant items are selected?
    \begin{equation}
        Precision = \frac{TP}{TP + FP}
    \end{equation}
    \begin{equation}
        Recall = \frac{ TP}{TP + FN}
    \end{equation}
    
    \item \textbf{F-scores (F1):} F1 score is simply the harmonic mean of precision and recall values.
    \begin{equation}
        F1 = 2*\frac{Precision * Recall}{Precision + Recall}
    \end{equation}
    
    \item \textbf{Area Under Curve (AUC):} Also known as "Area Under the Receiver Operating Characteristic curve (AUROC)." It is used to observe the usefulness of a model. AUROC curve is plotted between the true positive rate and the false positive rate at different threshold values.
    
\end{enumerate}

\subsection{Experiment I \& II: Result \& Analysis}
The first \& second experiment involves the use of Multi-layer feed-forward networks to model Credit Card Fraud data and then to use the proposed deep auto-encoder to classify the fraudulent patterns with greater accuracy. \par
To perform this, TensorFlow software library is used. It is an open-source library used for data-flow programming across a range of tasks.\\ \\
Test AUC Score: 95.33\%

\begin{figure}[ht]
	    \centering
	    \includegraphics[width=7.2cm]{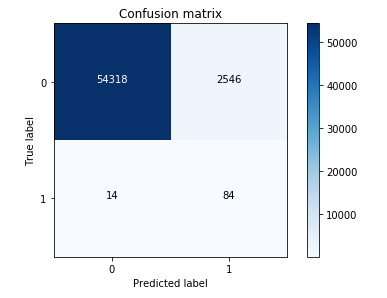}
	    \caption{Auto Encoder confusion matrix}
	    \label{pre1}
\end{figure}

\begin{figure}[ht]
	    \centering
	    \includegraphics[width=9cm]{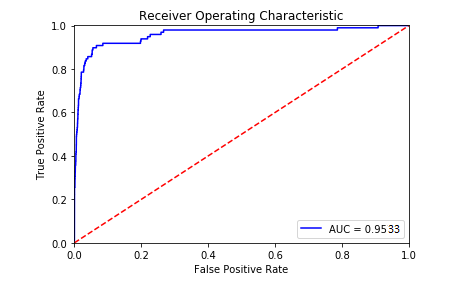}
	    \caption{Auto Encoder AUROC Curve}
	    \label{pre2}
\end{figure}

\begin{figure}[ht]
	    \centering
	    \includegraphics[width=9cm]{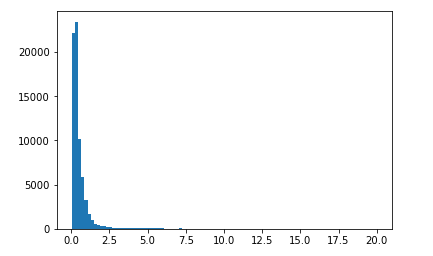}
	    \caption{Show distribution of all MSE}
	    \label{pre3}
\end{figure}

\begin{figure}[ht]
	    \centering
	    \includegraphics[width=9cm]{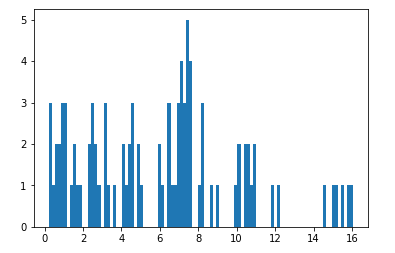}
	    \caption{Display only fraud cases}
	    \label{pre4}
\end{figure}

\clearpage
\subsection{Experiment III: Result \& Analysis}
The third experiment involves Nature Inspired Bat Algorithm for optimizing the FDS implemented using the first and the second experiment. The optimization is done using binary bat algorithm in the feature selection stage. It is found that this optimization can mitigate the training cost and the complexity of the entire system. Also it enhances the test AUC score significantly.\\ \\
Test AUC Score: 96.21\%

\begin{figure}[ht]
	    \centering
	    \includegraphics[width=9cm]{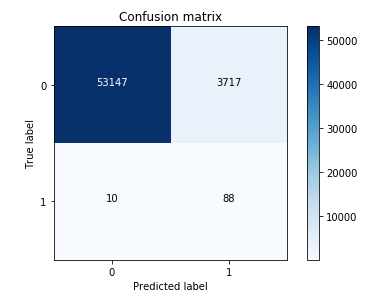}
	    \caption{Auto Encoder confusion matrix}
	    \label{pre5}
\end{figure}

\begin{figure}[ht]
	    \centering
	    \includegraphics[width=9cm]{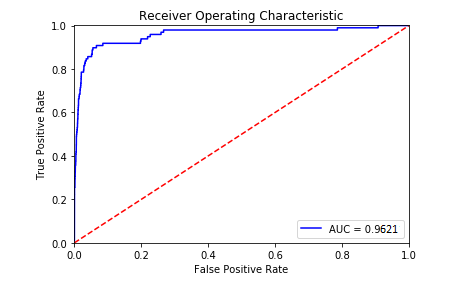}
	    \caption{Auto Encoder AUROC Curve}
	    \label{pre6}
\end{figure}

\newpage
\subsection{Experiment IV: Result \& Analysis}
The forth and final experiment involves Comparing different \textit{Scikit-learn} methods/algorithms to classify fraudulent patterns. These comparisons are done using different performance measures e.g.: \textit{AUC scores, confusion matrices and precision-recall curves}.

\subsubsection{FDS using different Scikit learn methods with Under-sampling}
This experimentation uses Under-sampling to handle class imbalance problem.\par
Under-sampling intends to balance class distribution by randomly eliminating majority class observations. This is practised until the majority and minority class instances are balanced out.
\begin{figure}[ht]
	    \includegraphics[width=9cm]{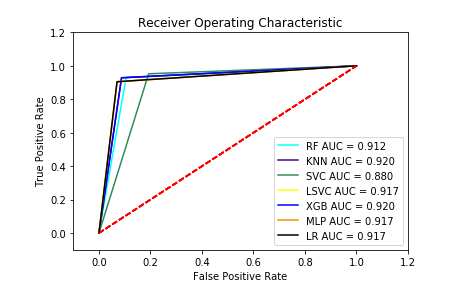}
	    \caption{AUCROC Curve with Under-sampling}
	    \label{pre7}
\end{figure}
    
We can see that XGB performs better than any other Scikit learn algorithms but the proposed FDS and the optimization outperforms it.

\subsubsection{FDS using different Scikit learn methods with Over-sampling}
This experimentation uses Synthetic Minority Over-sampling Technique (SMOTE) to handle class imbalance problem.\par
SMOTE is used to avoid over-fitting which occurs when exact replicas of minority class instances are added to the main data-set. \par 
A subset of data is taken from the minority class distribution as an example and then new, synthetic similar instances are created. These synthetic instances are then added to the original data-set. The new data-set is used as a sample to train the classification models. \\

\begin{figure}[ht]
	    \includegraphics[width=9cm]{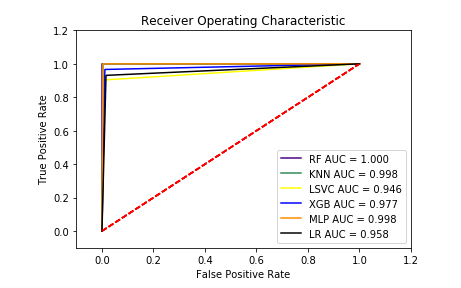}
	    \caption{AUCROC Curve with Over-sampling}
	    \label{pre8}
\end{figure}

\newpage
In this case Random Forest is the brightest performer, in fact it turns out to be almost an ideal classifier. But it might be the case that after applying SMOTE, the model gets over-fitted that means its credibility is not guaranteed.

\subsubsection{FDS using different Scikit learn methods with Stratified 3-fold sampling}
This experimentation uses different Scikit learn method to implement FDS on the given data-set. Also in this experiment \textit{Stratified 3-fold sampling} is used to demonstrate the different performance measures of these algorithms at three different folds. \\
Below is the best fold demonstration for different Scikit learn algorithms: 

\begin{itemize}
    \item \textbf{Random Forest (RF)}:
    \begin{figure}[!ht]
	    \centering
	    \includegraphics[width=9cm]{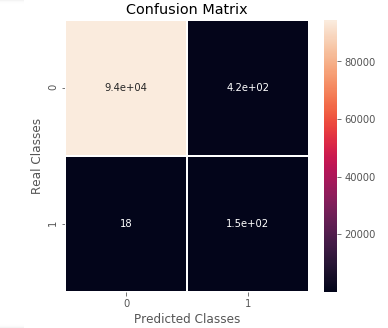}
	    \caption{RF Confusion Matrix}
	    \label{pre9}
    \end{figure}

    \begin{figure}[!ht]
	    \centering
 	    \includegraphics[width=9cm]{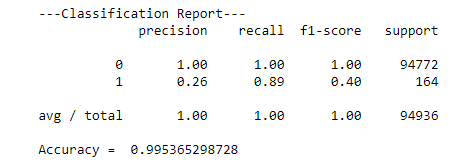}
	    \caption{RF Classification Report}
	    \label{pre10}
    \end{figure}
    
    \newpage
    \item \textbf{Linear Regression (LR)}:
    \begin{figure}[!ht]
	    \centering
	    \includegraphics[width=9cm]{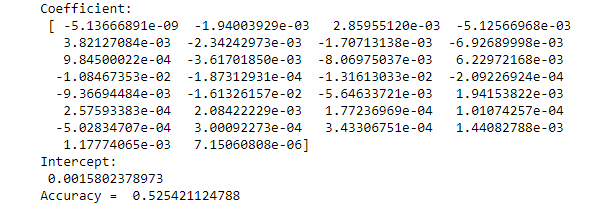} 
	    \caption{LR Classification Report}
	    \label{pre11}
    \end{figure}
    
    \item \textbf{Logistic Regression (LOR)}:
    \begin{figure}[!ht]
	    \centering
	    \includegraphics[width=9cm]{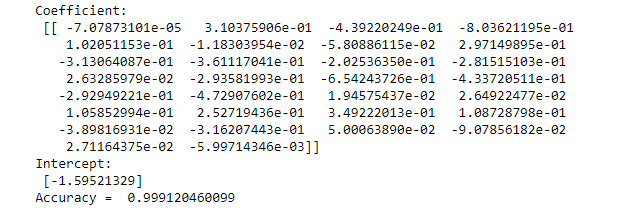}
	    \caption{LOR Classification Report}
	    \label{pre12}
    \end{figure}
    
    \item \textbf{Decision Tree Classifier using entropy criterion (DT)}:
    
    \begin{figure}[!ht]
	    \centering
	    \includegraphics[width=9cm]{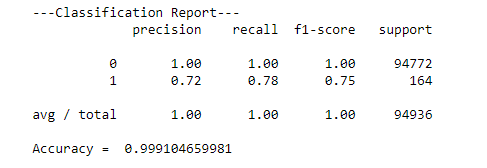}
	    \caption{DT with entropy Classification Report}
	    \label{pre13}
    \end{figure}
    
    \begin{figure}[!ht]
	    \centering
	    \includegraphics[width=9cm]{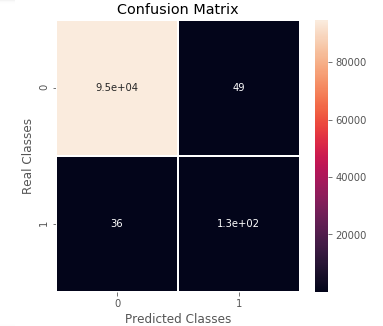}
	    \caption{DT with entropy Confusion Matrix}
	    \label{pre14}
    \end{figure}
    
    \newpage
    \item \textbf{Decision Tree Classifier using gini criterion (DT)}:
    \begin{figure}[!ht]
	    \centering
	    \includegraphics[width=9cm]{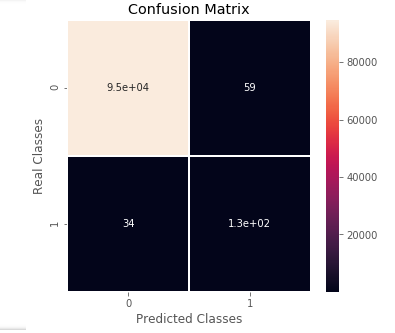}
	    \caption{DT with gini Confusion Matrix}
	    \label{pre15}
    \end{figure}
    
    \begin{figure}[!ht]
	    \centering
	    \includegraphics[width=9cm]{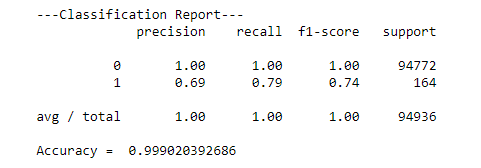}
	    \caption{DT with gini Classification Report}
	    \label{pre16}
    \end{figure}
    
    \newpage
    \item \textbf{Gradient Boosting Machine (GBM)}:
    \begin{figure}[!ht]
	    \centering
	    \includegraphics[width=9cm]{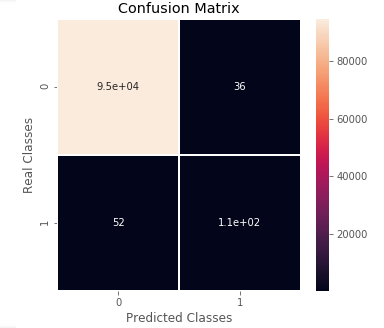}
	    \caption{GBM Confusion Matrix}
	    \label{pre17}
    \end{figure}
    \begin{figure}[!ht]
	    \centering
	    \includegraphics[width=9cm]{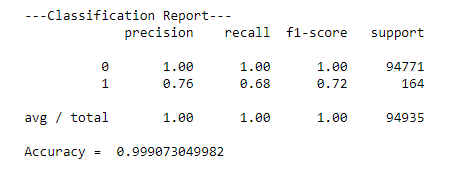}
	    \caption{GBM Classification Report}
	    \label{pre18}
    \end{figure}
    
    \newpage
    \item \textbf{XGBoost Classifier (XGB)}:
    \begin{figure}[ht]
	    \centering
	    \includegraphics[width=9cm]{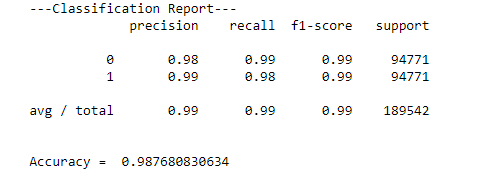}
	    \caption{XGBoost Classification Report}
	    \label{pre19}
    \end{figure}
    
    \item \textbf{ADABoost Classifier (ADAB)}:
    \begin{figure}[ht]
	    \centering
	    \includegraphics[width=9cm]{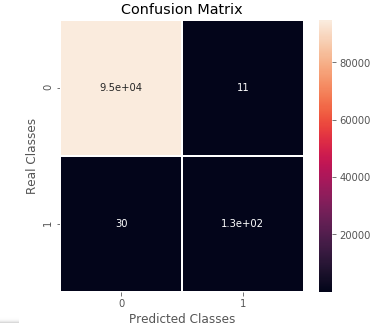}
	    \caption{ADABoost Confusion Matrix}
	    \label{pre20}
    \end{figure}
    \begin{figure}[ht]
	    \centering
	    \includegraphics[width=9cm]{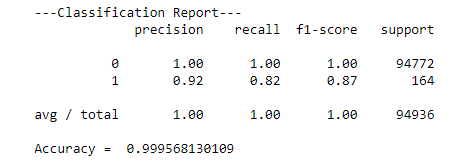}
	    \caption{ADABoost Classification Report}
	    \label{pre21}
    \end{figure}

\end{itemize}

In this case again Random Forest is the brightest performer, as we can see its false positive rate. But looking at the recall value, its below our expectations. i.e we can again say that out proposed FDS model with the optimization outperforms this case as well.

\newpage
\section{Conclusion}
The proposed methodology has made a useful contribution through the unsupervised
Fraud detection method. In fact this can be a step towards more automation in the Fraud detection systems. This is important because it reduces human intervention in the whole process and therefore mitigates time and cost. Also it obtains a high fraud coverage with low false alarm rate. \par
The supervised Fraud detection approaches have been shown to be an effective classifier. But the proposed method is useful even when labeled data is not available. \par
The proposed Fraud detection approach can be a very effective method especially
as the two fields (large amount of data and Deep learning) are rapidly evolving but more importantly because Deep learning is posited to be the most promising Machine learning method for Big Data analytic.






%




\bibliographystyle{IEEEtran}
\bibliography{bare_conf_compsoc}

\end{document}